\documentclass[10pt,twocolumn,letterpaper]{article}

\usepackage{cvpr}              

\usepackage[dvipsnames]{xcolor}

\usepackage{microtype}
\usepackage[accsupp]{axessibility}
\usepackage{graphicx}
\usepackage{caption}
\usepackage{subcaption} 
\usepackage{amsmath}
\usepackage{url}
\usepackage{wrapfig}
\usepackage[misc]{ifsym}

\usepackage[utf8]{inputenc} 
\usepackage[T1]{fontenc}    
\usepackage{url}            
\usepackage{booktabs}       
\usepackage{amsfonts}       
\usepackage{nicefrac}       
\usepackage{microtype}      
\usepackage{xcolor}         
\usepackage{pifont}
\usepackage{tabularx}
\usepackage{bm}
\usepackage{tablefootnote}
\usepackage[symbol]{footmisc}

\newcommand{\cmark}{\text{\ding{51}}}
\newcommand{\xmark}{\text{\ding{55}}}
\renewcommand{\paragraph}[1]{\textbf{#1. }}

\usepackage{xspace}
\def\modelname{msGFM\xspace}
\def\datasetname{GeoPile-2\xspace}

\definecolor{cvprblue}{rgb}{0.21,0.49,0.74}
\usepackage[pagebackref,breaklinks,colorlinks,citecolor=cvprblue]{hyperref}

\def\paperID{9014} 
\def\confName{CVPR}
\def\confYear{2024}

\title{Bridging Remote Sensors with Multisensor Geospatial Foundation Models}

\author{
Boran Han$^{1}$ \Letter
\quad
Shuai Zhang$^{1}$
\quad
Xingjian Shi$^{2}$ \thanks{Work done while at Amazon Web Services}
\quad
Markus Reichstein$^{1, 3}$\\
$^{1}$ Amazon Web Services \quad $^{2}$ Boson AI \quad $^{3}$ Max-Planck-Institute for Biogeochemistry
}

\begin{document}
\maketitle
\def\thefootnote{\Letter}\footnotetext{ \xspace Corresponding author}\def\thefootnote{\arabic{footnote}}

\begin{abstract}

In the realm of geospatial analysis, the diversity of remote sensors, encompassing both optical and microwave technologies, offers a wealth of distinct observational capabilities. Recognizing this, we present \modelname, a multisensor geospatial foundation model that effectively unifies data from four key sensor modalities. This integration spans an expansive dataset of two million multisensor images. \modelname is uniquely adept at handling both paired and unpaired sensor data. For data originating from identical geolocations, our model employs an innovative cross-sensor pretraining approach in masked image modeling, enabling the synthesis of joint representations from diverse sensors.  \modelname, incorporating four remote sensors, upholds strong performance, forming a comprehensive model adaptable to various sensor types. \modelname has demonstrated enhanced proficiency in a range of both single-sensor and multisensor downstream tasks. These include scene classification, segmentation, cloud removal, and pan-sharpening. A key discovery of our research is that representations derived from natural images are not always compatible with the distinct characteristics of geospatial remote sensors, underscoring the limitations of existing representations in this field. Our work can serve as a guide for developing multisensor geospatial pretraining models, paving the way for more advanced geospatial capabilities. Code can be found at \url{https://github.com/boranhan/Geospatial_Foundation_Models}

\end{abstract}  
\section{Introduction}
\label{sec:introduction}

\begin{figure*}[!t]
\centering
    \includegraphics[width=0.95\textwidth]{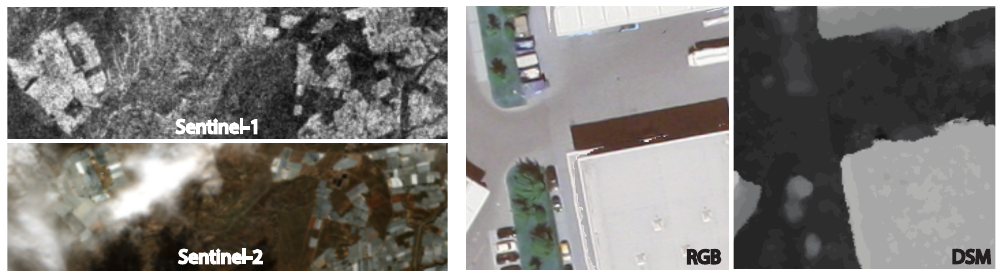}
    \caption{Examples of four sensor modalities: SAR, Sentinel 2, RGB, DSM. Here, each pair of \{SAR \& Sentinel-2\} and \{RGB \& DSM\} are colocated on the same geolocation respectively. In the example of Sentinel-2, only blue, green, red bands are shown for the convenience of visualization.}
    \label{fig:examples}
\end{figure*}

Geospatial remote sensors exhibit considerable diversity (Figure \ref{fig:examples}), with reported spatial \citep{spatialheterogeneity} and feature heterogeneity \citep{VANDERHOOF2023113498, rs14164083}.
Two principal categories emerge based on their imaging mechanisms: optical sensors (e.g., Sentinel-2 \citep{sentinel} and LiDAR) and microwave sensors (e.g., Synthetic-aperture radar \citep{SAR}). These sensors vary significantly in their observation methods and capabilities. Optical remote sensing captures reflected and absorbed electromagnetic radiation in the visible and near-infrared spectrum, yielding high-resolution imagery and surface property information. Conversely, microwave remote sensing operates at longer wavelengths, penetrating clouds and vegetation to reveal subsurface features and structural properties \citep{satellite_SAR} (Figure \ref{fig:examples}). 

A multisensor fusion approach combines the strengths of both optical and microwave remote sensing, offering a more comprehensive and accurate understanding of the Earth's surface \citep{Schmitt2017FusionOS}. By integrating data from multiple sensors, researchers can leverage the complementary nature of optical and microwave data to overcome limitations and obtain a more complete picture. For instance, combining optical and microwave data can help estimate soil moisture content, which is crucial for ecosystem management \citep{waterCO2, Agricultural}. Multisensor fusion also enhances the accuracy of topographic mapping by incorporating both surface features captured by optical sensors and elevation information derived from microwave sensors. Numerous multisensor fusion deep learning models have been proposed for individual tasks, such as cloud removal \citep{GLF-CR, MultisourceFusion}, biomass estimation \citep{AGB} and landcover segmentation \cite{ Contrastive_Multiview, MDAS}. These studies substantiate the enhancement in performance achievable by geospatial models incorporating multisensor modalities.

Despite these important synergies, most geospatial pretrained models focus predominantly on a single modality \citep{mendieta2023gfm, millionAID_supervised_pretraining, satmae, seco, ringmo}.  
While studies like \citet{multisource}, \citet{9614157} and \citet{9857009} employ Sentinel-2 and SAR for pretraining via contrastive learning, these methodologies are inherently limited by the need to paired sensor modalities. This limitation restricts the efficient utilization of the abundant unpaired sensor modalities that are prevalently available in real-world scenarios. By establishing a multisensor pretrained model scalable to both paired and unpaired sensors, a unified framework for analyzing multisensor remote sensing data can be provided. Such a model can be fine-tuned or used as a feature extractor to interpret multisensor data effectively. 

Therefore, our paper develops a novel multisensor geospatial pretraining model that can leverage many sensor modalities, paired or not. 
Additionally, our paper seeks to address several unexplored questions in the realm of multisensor geospatial models. A natural inquiry arises: \textit{How can joint representations between corresponding sensors be learned by employing masked image modeling techniques?} In geospatial tasks within the RGB domain, it is typical to leverage pretrained backbones on ImageNet \citep{DBLP:journals/corr/abs-2111-03690, wang2022advancing} or to utilize features distilled from such models \citep{mendieta2023gfm}. Given this, we inquire, \textit{Does leveraging or distilling features from established vision models enhance multisensor geospatial pretraining?} Lastly, a practical concern emerges: \textit{How can multisensor heterogeneity be mitigated during pretraining?}
Addressing these challenges is crucial for developing geospatial pretrained models capable of handling diverse sensor data. 
Our contributions can be summarized as follows:
\begin{itemize}
\item We introduce a novel cross-sensor paradigm, \modelname, for joint representation learning. This paradigm harmonizes diverse representations and empowers multisensor models to effectively discern the complex relationships between corresponding sensors.
\item We introduce a high-performing pretrained model, cultivated from a comprehensive multisensor pretraining dataset encompassing over 2 million images. This model adeptly amalgamates four sensor modalities: RGB images, Sentinel-2, SAR, and DSM, demonstrating superior performance across several important downstream tasks.
\item We demonstrate the synergistic advantages of incorporating multiple sensor modalities in pretraining, as opposed to focusing on single-sensor approaches. In addition, our work includes a thorough analysis of the model, offering practical insights and strategies for achieving optimal performance in multisensor foundation models.  
\end{itemize}

\begin{figure*}[h]
    \centering
    \includegraphics[width=1\textwidth]{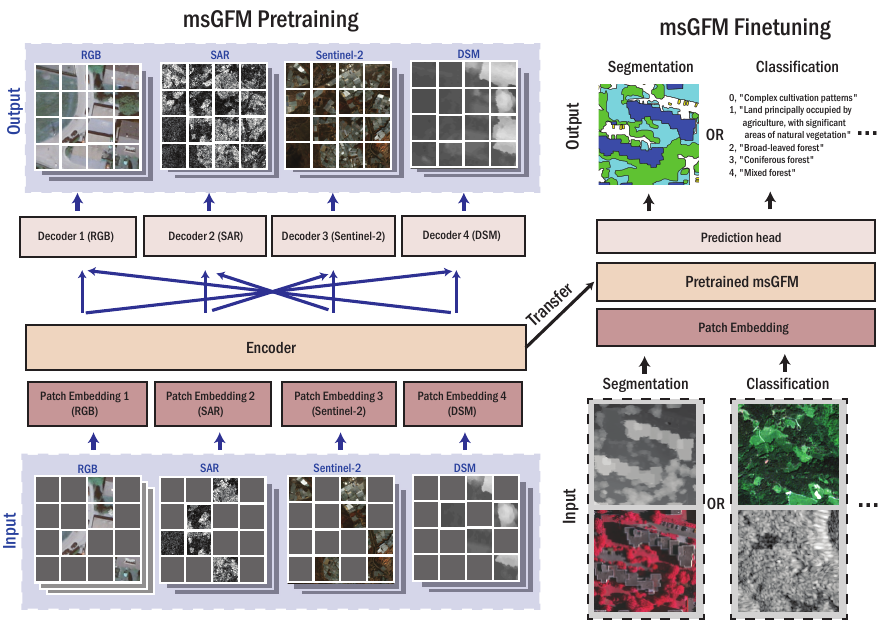}
    \caption{Overview diagram of \modelname. Each sensor is fed through a separate patch embedding layer (Section \ref{sec:input}) and through the same encoder. For reconstruction, separate decoders are used. If the sensors are paired, there's a chance that our model will reconstruct the corresponding paired sensor instead of itself (Section \ref{sec:cross}). Other best practices can be found in Section \ref{sec:moe}. In the finetuning stage, the pretrained encoder (\modelname) is transferred to different downstream applications with different prediction heads. In Appendix \ref{experimental_settings}, we discuss the usage of patch embedding in the downstream finetuning.}
    \label{fig:overview}
\end{figure*}

\section{Related Work}
\label{sec:related_work}

\textbf{Geospatial pretraining.} Geospatial technologies are becoming increasingly essential for applications, such as planning, monitoring and disaster response \cite{DLnature,Han_2021_ICCV, NEURIPS2023_f82ba6a6}. As pretrained models continue to revolutionize the fields of vision and language, their potential in the geospatial sphere is becoming increasingly evident. These models have demonstrated remarkable prowess in enhancing the efficacy of deep learning models when applied to downstream tasks \citep{indomain, seco, satmae, gassl, mendieta2023gfm}. 
The geospatial domain has seen the emergence of two main approaches for self-supervised pretraining paradigms. The first centers around the use of contrastive learning \citep{seco, gassl, multisource}. In this technique, the loss function is crafted to incentivize the model to draw similar or positive pairs closer together in the embedding space while pushing dissimilar or negative pairs further apart \citep{mocov2}. However, identifying appropriate augmentations for contrastive methods presents a significant challenge. Certain augmentations in geospatial images, which significantly alter the image's intensity, can lead to undesirable outcomes \citep{indomain}. Various implementations of pretraining with contrastive learning incorporate temporal and spectral augmentation \citep{seco}, while others apply a colorization objective \citep{colorOutofSpace}. Although works such as \citet{multisource}, \citet{9614157} and \citet{9857009} treat colocalized Sentinel-2 and SAR as positive pairs, these approaches are restricted to these two or more pairing sensor modalities and doesn't efficiently leverage the wide range of unpaired sensor modalities. Given these augmentation constraints \citep{indomain}, alternative methods have been developed, employing Masked Image Modeling (MIM) \citep{satmae, mendieta2023gfm, ringmo}, relying on simple spatial augmentations such as flipping and cropping. MIM not only requires less stringent augmentations but also outperforms its contrastive learning counterparts \citep{mendieta2023gfm, satmae, ringmo}. However, most prior studies focus on remote sensing imagery in the visible spectrum or employ a single sensor modality \citep{mendieta2023gfm, millionAID_supervised_pretraining, satmae, seco}. Alternatively, they are confined to \textit{two or more paired} sensors due to the inherent limitations of contrastive learning \citep{multisource, 9614157, 9857009}. In this work, we develop our pretraining objective based on the masked image modeling approach, akin to \citep{simmim, mae}. We demonstrate that our model can be pretrained with four sensor modalities, taking advantage of the unpaired sensor.


\textbf{Multi-source learning in language and vision communities.} Multi-source learning is a prevalent strategy when handling multi-modal \citep{shen2023scaling, Uni-Perceiver-MoE} and multitask challenges \citep{chen2023lightweight, chen2024camml} in both the language and vision domains \citep{modsquad, Aoki_2022, m3vit, muppet, xlm, xlmr, multimae}. This technique exploits data from diverse sources to bolster the learning process and enhance model performance. A notable example is multilingual pretrained models, such as XLM \citep{xlm} and its derivatives \citep{xlmr, xlme}. These models utilize multilingual datasets, pretraining them on a large scale to generate unsupervised cross-lingual representations \citep{xlmr, xlm}. This approach enables the models to develop a unified representation across multiple languages, thereby enhancing their performance on cross-lingual tasks. Furthermore, the batching strategy has been identified as an essential aspect of creating generalizable representations and preventing collapse in multilingual models \citep{muppet, aghajanyan2020better}. Simultaneously, the Mixture-of-Experts (MoE) strategy \citep{shazeer2017outrageously} has been utilized to enhance multi-source learning in both multitask learning \citep{modsquad, Aoki_2022, m3vit} and language-vision pretraining \citep{shen2023scaling, Uni-Perceiver-MoE}. In the specific context of multisensor geospatial pretraining, heterogeneity can originate from the use of different sensor types (e.g., optical, microwave) or different platforms (e.g., various satellites). Properly addressing this heterogeneity is crucial as it can significantly influence the performance of the pretraining model \citep{muppet}. To meet this challenge, we draw inspiration from works in vision \citep{riquelme2021scaling}, language \citep{muppet, xlm} and vision-language models \cite{Uni-Perceiver-MoE}. We incorporate techniques such as cross-sensor representation learning into our \modelname.

\section{Cross-sensor geospatial pretaining}
\label{sec:model}

In this section, we present the multisensor pretraining paradigm. Following \citep{mendieta2023gfm}, we employ SIMMIM \citep{simmim} using a Swin Transformer \citep{swin, liu2021swinv2} as a backbone. Figure \ref{fig:overview} presents an overview of the cross-sensor geospatial pretraining methodology.

\subsection{Input representation}
\label{sec:input}

\textbf{Distinct embedding layers for each sensor.} We consider $N$ total types of sensor modalities, with each sensor having a corresponding number of channels, denoted as $\{C_i\}_{i=1...N}$. Taking into account the unique number of channels associated with each sensor (some examples shown in Table \ref{tab:pretraining}), we utilize individual patch embeddings tailored to each specific sensor. This approach allows the model to efficiently process and learn from the distinct characteristics of various sensor modalities.

The patch embedding is primarily obtained from convolution layers. To elaborate, for the image from the $i$-th sensor, $\bm{I} \in \mathbb{R}^{W \times H \times C_i}$. Here, $W$ and $H$ represent the width and height of the images. We first apply the convolution layers, $\{f_i\}_{i=1...N}: \mathbb{R}^{W \times H \times C_i} \rightarrow \mathbb{R}^{W \times H \times C_e}$. The function $f_{i}$ represents the convolution neural network layers for images from the $i$-th sensor. In our work, $C_e$ is the same for all sensors. Subsequently, the output is segmented into square patches of size $P$, yielding $\bm{T}_i \in \mathbb{R}^{L \times P^2 C_e}$, where $L$ represents the total number of patches. To effectively manage the channel heterogeneity inherent in different sensor modalities, each sensor modality is processed through its own trainable embedding layer. This step standardizes the representation dimensions before they are input into the shared encoder. 

\subsection{Cross-sensor pretraining}
\label{sec:cross}

\textbf{Shared encoder for all sensor modalities. } The patches obtained from $\{f_i\}_{i=1...N}$, $\bm{T}_i \in \mathbb{R}^{L \times P^2 C_e}$, will then be masked and fed through the encoder. The masking strategy employed in our approach is the same as those used in \cite{simmim}. By having separate patch embedding layers ($f_{i}$) for each sensor, the model can learn the unique characteristics of each sensor modality. The learned embeddings from all sensors are then integrated through the same encoder, enabling the model to effectively learn joint representations and handle multisensor geospatial data.

\textbf{Separate decoder for each sensor and cross sensor prediction.} Collecting data from different sensors for the same geo-location is a common practice in the geospatial domain. Learning joint representations of such multisensor data can prove beneficial for various downstream tasks. Although contrastive learning has demonstrated promise in learning effective representations, its performance may be limited due to the lack of suitable data augmentations for remote sensing images \cite{indomain}. To address this issue, we propose employing cross-sensor strategies in the context of MIM to learn joint representations for multisensor geospatial data. For instance, when the model is fed with masked images from DSM, it can predict the masked patches of itself or the corresponding images from RGB. An example pair of DSM and RGB images is shown in Figure \ref{fig:examples} in the two panels on the right. This encourages the model to align the different sensor representations. Accordingly, our model incorporates different decoders for each sensor.

Specifically, if there exists a pair of images from the $i$-th sensor and $j$-th sensor, $\{\bm{I}_i \in \mathbb{R}^{W \times H \times C_i}, \bm{I}_j \in \mathbb{R}^{W \times H \times C_j}\}$, the model processes the masked image as follows:
\begin{align}
\centering
\bm{I}'_i &= D_i(En(f_i(\bm{I}_i))) \text{ , or } \bm{I}'_j = D_j(En(f_i(\bm{I}_i))) \nonumber \\
\text{and } \bm{I}'_j &= D_j(En(f_j(\bm{I}_j))) \text{ , or } \bm{I}'_i = D_i(En(f_j(\bm{I}_j)))
\label{eq:cross_sensor}
\end{align}
where $En: \mathbb{R}^{L \times P^2 C_e} \rightarrow \mathbb{R}^{L \times C_m}$ is the shared encoder, and $C_m$ is the embedding dimension of the final layer in the encoder. $I'_i$ and $I'_j$ are the predicted $i$-th and $j$-th sensor type respectively. $D_i: \mathbb{R}^{L \times C_m} \rightarrow \mathbb{R}^{W \times H \times C_i}$ is the decoder to reconstruct the $i$-th sensor type. Equation \ref{eq:cross_sensor} shows that the predicted output of the pretraining model will either reconstruct itself or its paired sensor images. 

If there's no paired sensor in the pretraining dataset, it will construct itself in the conventional way: 
\begin{align}
\centering
\bm{I}'_i = D_i(En(f_i(\bm{I}_i)))
\label{eq:single_sensor}
\end{align}

This approach capitalizes on the inherent relationship between different sensors observing the same location, enabling the model to capture complementary information. Furthermore, it provides flexibility in handling scenarios where no paired sensors are available, allowing for enhanced adaptability in choosing the pretraining dataset. This is particularly advantageous given that multisensor geospatial datasets are less prevalent than single-sensor datasets.

\subsection{Pretraining data.} 

\begin{table*}[h!]

    \centering
    \setlength\tabcolsep{5.0pt} 
    \small
    \begin{tabular}{cccccc}
        \toprule
        Dataset & \# Images & GSD &  Sensor modality & \# Channels & paired sensors?\\
        \toprule
        GeoPile \citep{mendieta2023gfm} & 600K & 0.1m - 30m & RGB$^a$ & 3 & \xmark\\
        MillionAID \citep{Long2021DiRS}  & 1M & 0.5 - 153m & RGB$^a$ & 3 & \xmark\\
        SEN12MS \citep{sen12ms} & 320K & 10m & SAR / sentinel-2 & 2/14 & \cmark\\
        MDAS \citep{MDAS} & 40K & 0.1m - 10m & DSM/ RGB$^b$ & 1/3 & \cmark\\
        \midrule
    \end{tabular}
    \caption{Breakdown of datasets of our pretraining data. We gather approximately 2M samples from a combination of labeled and unlabeled satellite imagery with various ground sample distances and sensor modalities. GeoPile and MillionAID are not sourced from a single sensor; instead, they amalgamate sensor images from an array of satellites, including NAIP, GeoEye, WorldView, QuickBird, IKONOS, and SPOT satellites, among others. MDAS is derived from airborne sources \citep{MDAS}. For more in-depth details regarding the RGB, please refer to Appendix \ref{wo_RGB_modality}}
    \label{tab:pretraining}
\end{table*}

Our multisensor pretraining data, \datasetname, is composed of four sensor modalities, amassed to a total of 2 million images through the inclusion of additional geospatial data. The detailed composition of \datasetname is presented in Table \ref{tab:pretraining}. Specifically, \datasetname incorporates SEN12MS \citep{sen12ms}, a dataset enriched with paired SAR and Sentinel-2 satellite images from all meteorological seasons, to augment data diversity. All sensors in this dataset are ortho-rectified \citep{sen12ms}. Additionally, we have integrated DSM and RGB images from the MDAS dataset \citep{MDAS}, resized to a dimension of 384.

It is important to note that while the Sentinel-2 modality includes RGB channels as part of its imaging band, we have distinguished and separated this RGB modality due to its extensive dataset that surpasses the scope of Sentinel-2. This dataset exhibits a wide range of Ground Sample Distances (GSD) and high feature entropy. These attributes, beyond just the imaging band, have been proven to be influential in pretraining, as evidenced in studies like \citep{mendieta2023gfm, satmae}. Our observations indicate that excluding the RGB modality from our pretraining dataset (\datasetname) leads to a decrease in effectiveness, as opposed to when it is included (see Appendix \ref{wo_RGB_modality}). An enhanced version of \datasetname, which includes RGB data from sources like GeoPile \citep{mendieta2023gfm} and MillionAID \citep{Long2021DiRS}, shows improved performance across the seven downstream tasks specified in GFM \citep{mendieta2023gfm}, under similar experimental conditions (detailed in Appendix \ref{pretraining_data}). Further optimization of \datasetname-RGB was achieved through experiments with various datasets (refer to Appendix \ref{pretraining_data}).

\subsection{Best Practices in Pretraining}
\label{sec:moe}

A shared encoder can present challenges when it comes to efficiently learning each sensor's representation. To tackle this issue, we propose integrating the sparsely gated Mixture of Experts (MoE) approach \citep{shazeer2017outrageously} to replace MLP layers within the encoder. Our pretraining loss function, $L$, combines L1 loss \citep{simmim, mae} for reconstruction (i.e., MIM loss) and auxiliary losses \citep{tutel, riquelme2021scaling}: $L = L_\text{MIM} + \lambda L_\text{auxiliary}$, where $\lambda$ represents the weight for auxiliary losses. In practice, we use $\lambda = 0.01$. 

We sequentially load all sensor data in our model. This approach ensures that our model's optimization spans all tasks. Specifically, each batch in our model is constituted as a set: ${\bm{I} \in \mathbb{R}^{W \times H \times C_i}}_{i=1...N}$. Such a methodology is commonly utilized in multitask learning, aiming to forge a unified representation across diverse tasks during the training process, despite their distinct learning objectives. Our investigations reveal that this strategy is equally effective in the context of multisensor geospatial pretraining. Furthermore, considering the unique imaging mechanisms of these sensors, we choose to initiate pretraining \textit{from scratch}, as detailed in Section \ref{sec:fromscratch}.

\section{Experiments}
\label{sec:experiments}

\textbf{Experimental Settings.} All of our experiments are conducted using a Swin-base architecture \citep{liu2021swinv2} with a patch size of 16×16 pixels and 8 experts. The models are pretrained for either 100 epochs for ablation studies or 800 epochs to achieve optimal results and maintain comparability with state-of-the-art methods. When specified, 1\% BigEarthNet (BEN) \cite{BEN} and 1\% SEN12MS-CR are also employed for ablation studies. We utilize 8 NVIDIA V100 GPUs with a batch size of 2048 (128 per GPU) and an image size of 192$\times$192. All pretraining settings follow the configurations described in \citep{mendieta2023gfm}. Detailed pretraining settings and pretraining cross-sensor reconstruction visualization can be found at Appendix \ref{pretraining_settings} and Section \ref{visualizaion} respectively.

\textbf{Downstream Evaluation.} Upon completion of the pretraining, we fine-tune and assess the model on a diverse range of downstream multisensor datasets. This aims to provide a comprehensive understanding of the model's performance potential across various tasks. Table \ref{tab:downstream} provides an overview of the downstream evaluation tasks, together with their respective sensor modalities. Among these tasks, the use of multisensor data can enhance the performance of land classification and segmentation. Meanwhile, cloud removal is inherently dependent on multisensor modalities and cannot be effectively tackled without them. Although pansharpening requires one optical sensor, it relies heavily on multi-spectral images. 

\begin{table*}[h]
    
    \centering
    \setlength\tabcolsep{5.0pt} 
    \small
    \begin{tabular}{ccccc}
        \toprule
        Dataset & \# Application & GSD &  Sensor modality & \# Channels\\
        \toprule
        Big Earth Net \citep{BEN} & Scenes classification & 10m - 60m & SAR / Sentinel-2 & 2/14\\
        Vaihingen \citep{vaihingen} & Land segmentation & 0.09m & DSM / RGB & 1/3\\
        SEN12MS-CR \citep{sen12mscr} & cloud removal & 10 - 60m & SAR / Sentinel-2 & 2/14\\
        SpaceNet & Pan-sharpening & 0.1m - 10m & WorldView 3 & 8\\
        \midrule
    \end{tabular}
    \caption{Downstream tasks. It covers various use cases in geospatial domain, with a range of ground sample distances and sensor modalities.}
    \label{tab:downstream}
\end{table*}

\subsection{Visualizing reconstruction quality}
\label{visualizaion}

\begin{figure}[h!]
\centering
    \includegraphics[width=0.48\textwidth]{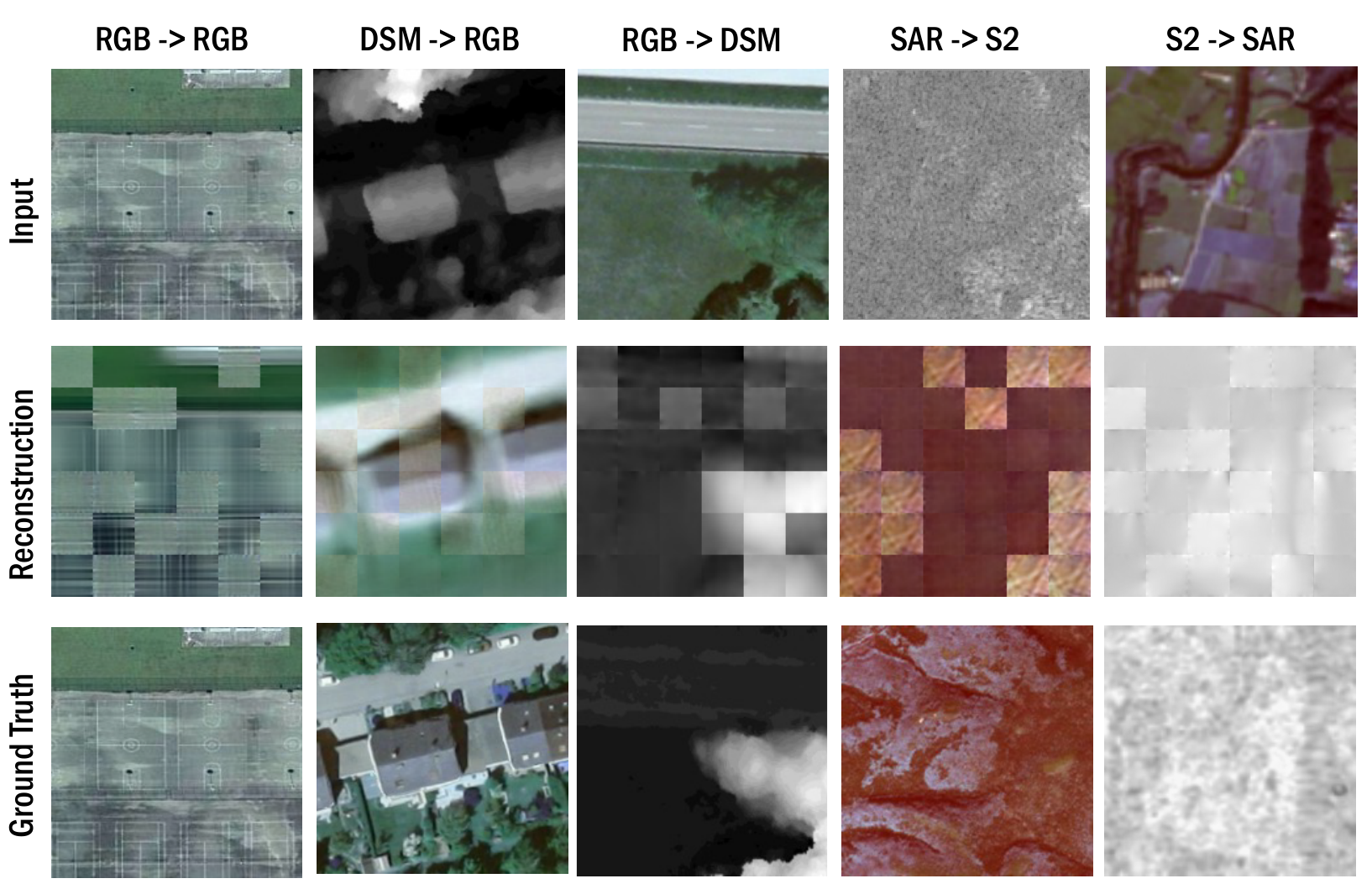}
    \caption{Examples of cross-sensor pretraining. The first row represents the input before masking, the second row depicts the reconstruction, and the third row shows the ground truth.}
    \label{fig:mim_examples}
\end{figure}

To demonstrate this methodology, we provide several examples in Figure \ref{fig:mim_examples}. These instances visually illustrate our cross-sensor pretraining approach, highlighting the ability of RGB to self-reconstruct, as well as the excellent cross reconstruction capabilities between DSM and RGB images. However, self-reconstruction and cross reconstruction involving SAR images pose some challenges, as we use unprocessed, noisy SAR images. Due to the structure of MIM, which involves an encoder and a lightweight reconstruction decoder, only low-frequency components of the images are reconstructed \cite{simmim, mae}, making the SAR reconstruction slightly difficult. Specifically, we visualized the statistics \citep{C_bandmicrowaves} and SSI \citep{Sheng1996ACE} value before and after the reconstruction for both HV and VV bands (Figure \ref{fig:SAR_backscatter}).

\subsection{Geospatial downstream evaluation}
\label{sec:comparison}
\begin{table*}[h]

    \centering
    \setlength\tabcolsep{5.0pt} 
    \small
    \begin{tabular}{cccccccccc}
        \toprule
        Methods & 10\% BEN & 100\% BEN & \multicolumn{3}{c}{SEN12MS-CR} &  \multicolumn{2}{c}{SpaceNet} & Vaihingen \\
         & mAP ($\uparrow$) & mAP ($\uparrow$) & MAE ($\downarrow$) & SAM ($\downarrow$) & SSIM ($\uparrow$) & PSNR ($\uparrow$) & SSIM ($\uparrow$) & mIOU ($\uparrow$) \\
        \toprule
        \toprule
        SeCo \citep{seco} & 82.6 & 87.8 & - & - & - & - & -  & 68.9 \\
        SatMAE \citep{satmae} & 82.1 & - & - & - & - & 22.742 & 0.621 & 70.6  \\
        MoCoV2 \citep{mocov2} & - & 89.3 & - & - & - & - & - & -\\
        DINO-MC \citep{dinomc} & 84.2 & 88.6 & - & - & - & - & - & - \\
        GFM \citep{mendieta2023gfm} & 86.3 & - & - & - & - & 22.599 & 0.638  & 75.2  \\
        Random & 82.6 & 86.2 &  0.048 & 14.78  & 0.572 & 21.825 & 0.594 & 67.0 \\
        IN-22k \citep{liu2021swinv2} & 85.7 & 89.5 & - & - & - & 21.655 & 0.612 & 74.7  \\
        \modelname & \textbf{87.5} & \textbf{92.9} & \textbf{0.026} & \textbf{4.87} & \textbf{0.842} & \textbf{22.850} & \textbf{0.668} & \textbf{75.8} \\
        \midrule
    \end{tabular}
    \caption{Quantitative results of all the downstream tasks (Table \ref{tab:downstream}) from \modelname (ours) compared to other pretrained models. 
    Results are replicated from the previous reports. Random: random initialization.}
    \label{tab:comparesota}
\end{table*}

\subsubsection{Scene classification.} 
One prevalent remote sensing application is classification. We evaluate our pretraining model on BEN, a dataset extensively used in other literature \citep{seco, satmae, mocov2, dinomc, mendieta2023gfm}. BEN \citep{BEN} is a large-scale imbalanced remote sensing dataset specifically designed for multi-label classification tasks. The data includes pairs of 12-band Sentinel-2 images along with their corresponding 2-band SAR images. We employ the data split and 19-class evaluation, as is standard in the literature \citep{indomain, seco, satmae,  mendieta2023gfm}. In Table \ref{tab:comparesota}, we report the mean average precision (mAP) results on BEN for all methods. Our model can provide robust performance against other pretraining methods \citep{seco, satmae, mocov2, dinomc, mendieta2023gfm, liu2021swinv2}, including ImageNet-22k \citep{liu2021swinv2}. 
More results of the random initialization and ImageNet initialization can found in previous studies \citep{mendieta2023gfm, satmae}. 

\begin{table}[h]
    
    \centering
    \begin{tabular}{ccccc}
        \toprule
        Methods & MAE ($\downarrow$) & SAM ($\downarrow$) & SSIM ($\uparrow$) \\
        \toprule
        \toprule
        SAR-Opt-cGAN \citep{SAR-Opt-cGAN} & 0.043 & 15.49 & 0.764 \\
        DSen2-CR \citep{DSen2-CR} & 0.031 & 9.47 & 0.874 \\
        GLF-CR \citep{GLF-CR} & 0.027 & 7.65 & \textbf{0.885}\\
        \modelname & \textbf{0.026} & \textbf{4.87} & 0.842\\
        \midrule
    \end{tabular}
    \caption{Quantitative results of cloud removal, compared to existing models that are specially designed for cloud removal. Results are replicated from the original paper.}
    \label{tab:cloud}
\end{table}

\subsubsection{Cloud removal}
The majority of optical observations acquired via spaceborne Earth imagery are affected by clouds, presenting challenges in reconstructing cloud-covered information in current studies. While optical imagery is impacted by adverse weather conditions and the lack of daylight, SAR sensors remain unaffected, offering a valuable source of complementary information. Consequently, performing cloud removal tasks without SAR data significantly degrades task performance \citep{GLF-CR}. We evaluate our model on SEN12MS-CR \citep{sen12mscr}. The results, presented in Table \ref{tab:cloud}, show promising performance in Spectral Angle Mapper (SAM) and Mean Absolute Error (MAE), outperforming existing cloud removal models \citep{SAR-Opt-cGAN, DSen2-CR, GLF-CR}. Additionally, the Structural Similarity Index Measure (SSIM) metric yields comparable results to these methods. Our multisensor pretraining approach, incorporating SAR data, facilitates effective cloud removal. In contrast, other geospatial pretraining models that rely solely on optical data fall short in demonstrating their cloud removal capabilities, as shown in Table \ref{tab:comparesota}.

\subsubsection{Pan-sharpening.} 
Pansharpening, akin to super-resolution, involves combining a high-resolution grayscale panchromatic image with the color information from a low-resolution multispectral image to generate a high-resolution color image. For this assessment, we utilized the SpaceNet2 dataset, following the methods in \citep{mendieta2023gfm}. We juxtaposed the performance of our model with a series of baseline models, measuring the outcomes using peak signal-to-noise ratio (PSNR) and SSIM. As illustrated in Table \ref{tab:comparesota}, our model demonstrates superior performance over its competitors. Notably, the SpaceNet dataset, which comprises images from the WorldView-3 satellite—a source not included in our pretraining data—demonstrates the strong transferability of our pretrained model across diverse sensors.

\subsubsection{Segmentation} 
Segmentation is another popular remote sensing application for enabling automated extraction of building footprints or land cover mappings over wide regions. We therefore conduct experiments on this task using Vaihingen \citep{vaihingen}, which is an urban semantic segmentation dataset collected over Vaihingen, Germany at a GSD of 0.9m. The experiment settings are the same as \citep{mendieta2023gfm}. We report the intersect of union (IoU) segmentation results for all methods in Table \ref{tab:comparesota}. Our approach is able to provide the best result. 

\subsection{Comparison with single sensor pretraining.}
\begin{figure}[!h]
\centering
	\includegraphics[width = 8.5cm]{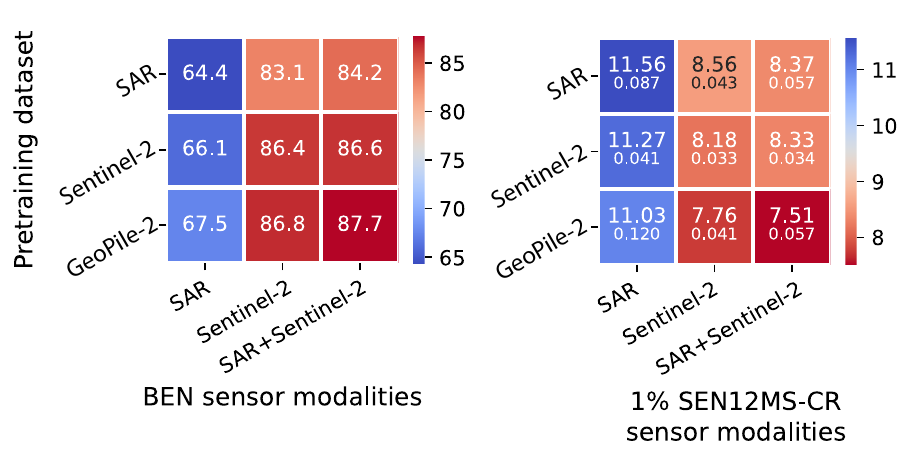}
\caption{Comparison of our multisensor approach with single modality pretraining on 10\% of the BEN dataset (left) using mAP ($\uparrow$) and 1\% of SEN12MS-CR (right) using SAM ($\downarrow$). Given the reduced dataset size for cloud removal (1\% of SEN12MS-CR), we conduct the experiment in three replicates and report both the mean (top line in each cell) and standard deviations (bottom line in each cell).}
	\label{fig:modalities}
\end{figure}

To underscore the pivotal role of multiple sensor modalities in pretraining and validate that our method leads to multisensor synergy, we compare our multisensor pretraining approach using \datasetname with models pretrained on only one sensor modality (i.e., either SAR or Sentinel-2) from SEN12MS \citep{sen12ms}. We assess the performance of these models on the BEN \citep{BEN} and SEN12MS-CR \citep{sen12mscr}, employing both sensors individually and in combination. Figure \ref{fig:modalities} highlights two advantages of our model: (1) The multisensor pretraining model consistently outperforms models pretrained with a single sensor modality, as indicated by superior performance across all columns when the sensor modality is fixed. (2) Using both sensors for tasks like land use classification and cloud removal leads to enhanced performance, demonstrated by higher accuracy across all rows when the pretraining data is fixed.

The second advantage can be attributed to the complementary data provided by both sensors. Sentinel-2 images are widely used for land use and land cover classification tasks due to their rich spectral information. In contrast, SAR images offer critical insights for identifying water bodies and urban structures, capturing the radar backscatter properties of the Earth's surface. We observe that different tasks have various responses to multisensor data, a phenomenon also evident in the cloud removal task previously studied \citep{GLF-CR}.

Notably, even when evaluating the BEN dataset using only the Sentinel-2 modality, our method still achieves superior results compared to other pretraining methods (86.8\%). This demonstrates that the improvement of \modelname is not solely due to an increase in sensor modality within the downstream tasks.

\subsection{Pretraining from scratch performs better.}
\label{sec:fromscratch}

\begin{table*}[h]

    \centering
    \setlength\tabcolsep{5.0pt} 
    \small
    \resizebox{\textwidth}{!}{
    \begin{tabular}{cccccccccc}
        \toprule
        Methods & 1\% BEN & 10\% BEN & \multicolumn{3}{c}{SEN12MS-CR} &  \multicolumn{2}{c}{SpaceNet} & Vaihingen \\
         & mAP ($\uparrow$) & mAP ($\uparrow$) & MAE ($\downarrow$) & SAM ($\downarrow$) & SSIM ($\uparrow$) & PSNR ($\uparrow$) & SSIM ($\uparrow$) & mIOU ($\uparrow$) \\
        \toprule
        \toprule
        Distilled from ImageNet22K \citep{imagenet} & 79.4 & 86.4 & 0.035 & 6.42 & 0.726 &22.107 & 0.621 &  72.9\\
        Distilled from CLIP \citep{clip} & 76.6 & 83.8 & 0.051 & 8.96 & 0.707 & 22.559 & 0.674 & 69.3 \\
        Reconstruct CLIP (EVA \citep{eva}) & 73.5 & 80.6 & 0.053 & 9.96 & 0.689 & 21.778 & 0.591 & 65.7\\
        From scratch & \textbf{80.9} & \textbf{87.2} & \textbf{0.026} & \textbf{5.04} & \textbf{0.821} & \textbf{22.742} & \textbf{0.677} & \textbf{74.8}\\
        \midrule
    \end{tabular}}
    \caption{Distillation from other pretraining model vs pretraining from scatch}
    \label{tab:geopile}
    \vspace{-0.0cm}
\end{table*}

\begin{table*}[h!]
    \centering
    \setlength\tabcolsep{5.0pt} 
    \small
    \resizebox{\textwidth}{!}{
    \begin{tabular}{ccccccccccc}
        \toprule
        \multicolumn{2}{c}{Pretraining strategies} & Cross sensor & 1\% BEN & 10\% BEN& \multicolumn{3}{c}{SEN12MS-CR} & \multicolumn{2}{c}{SpaceNet}  & Vaihingen\\
        MoE & cross-sensor& percentage &  mAP  ($\uparrow$) & mAP  ($\uparrow$) & MAE ($\downarrow$) & SAM ($\downarrow$) & SSIM ($\uparrow$) & PSNR ($\uparrow$) & SSIM ($\uparrow$) & mIOU ($\uparrow$) \\
        \toprule
        \toprule
        \xmark & \xmark & 0\% & 78.3 & 86.2 & 0.038 & 8.19 & 0.735 & 22.333 & 0.589 & 72.8\\
        \cmark & \xmark & 0\% & 78.5 & 86.2 & \textbf{0.026} & 5.11 & 0.767 & 22.528 & 0.637  & 73.4 \\
        \xmark & \cmark & 50\% &  \textbf{80.7} & 86.9 & 0.036 & 8.67 & 0.753 & 22.518 & 0.611 & \textbf{74.6} \\
        \cmark & \cmark & 100\% & 80.5 & 86.8 & \textbf{0.026} & \textbf{4.96}  & 0.789 & 22.634 & 0.649 & 74.4 \\
        \cmark & \cmark & 50\% & \textbf{80.9} & \textbf{87.5} & \textbf{0.026} & 5.04  & \textbf{0.821} & \textbf{22.742} & \textbf{0.677} & \textbf{74.8} \\
        \midrule
    \end{tabular}}
    \caption{Quantitative results of \modelname, with and without MoE/cross sensor reconstruction.}
    \label{tab:ablation}
\end{table*}

In geospatial tasks, especially within the RGB spectrum, it is common to use backbones pretrained on ImageNet \citep{DBLP:journals/corr/abs-2111-03690, wang2022advancing}, or to leverage features distilled from such models \citep{mendieta2023gfm}. Consequently, we assess the efficacy of leveraging established vision pretrained models for multisensor geospatial pretraining. For fair comparisons, all experiments are trained for 100 epochs.

In one experiment, we extract intermediate features following the methodology in \citep{mendieta2023gfm} and benchmark them against embeddings obtained from ImageNet-22k. We also conduct a parallel experiment utilizing the CLIP model \citep{clip}, known for its robust multimodal representation learning.

The results, shown in Table \ref{tab:geopile}, indicate that feature distillation from ImageNet-22k outperforms that from CLIP \citep{clip} in terms of performance. Additionally, we explore the EVA method \citep{eva}, which deviates from traditional MIM approaches like MAE \citep{mae} by reconstructing CLIP features of masked patches rather than the patches themselves. Contrary to expectations, EVA, despite its established superiority in other contexts, does not perform as well in our downstream evaluations. This indicates that CLIP features \citep{clip} may face a significant domain gap when applied to multisensor geospatial data.

Conversely, the highest accuracy for multisensor geospatial pretraining is achieved when models are trained from scratch. The lower performance of distillation methods is attributed to the pronounced domain gap between natural images and geospatial-specific sensors. Furthermore, distillation inherently limits the student model’s performance to that of the teacher model \citep{mendieta2023gfm}. This domain gap arises from fundamental differences in the physical mechanisms of optical and microwave remote sensing: while optical remote sensing relies on the reflection and absorption of electromagnetic radiation, microwave remote sensing operates based on the scattering, penetration, and dipole-interference of microwaves \citep{SAR}. Given that natural images are primarily captured by optical sensors, this leads to a considerable domain discrepancy. 

This finding highlights the need for robust foundation models tailored to the geospatial domain, capable of accommodating diverse sensor data and improving multisensor task performance.

\subsection{Ablation studies}
\label{sec:ablation}

In the proposed \modelname model, we incorporate both cross-sensor pretraining paradigms and the Mixture of Experts (MoE). In an ablation study, we present the results when either MoE or cross-sensor pretraining is omitted. As shown in Table \ref{tab:ablation}, removing MoE from the model results in similar performance on the some datasets, while other tasks see a more substantial decrease. This uneven response across different tasks aligns with observations made in several previous multi-modal studies \citep{Uni-Perceiver-MoE}. On the other hand, removing the cross-sensor paradigm leads to a consistent performance decline across all tasks.

It is natural to question how the proportion of sensor crossing affects performance. To explore this, we perform an ablation study on the percentage of sensors subject to cross-reconstruction. Our results suggest that a sensor crossing rate of 50\% provides slightly superior outcomes compared to a rate of 100\%. This indicates that the optimal sensor crossing strategy maintains a balance between the benefits of cross-reconstruction and the retention of sensor-specific information, consequently enhancing performance across a diverse range of geospatial tasks.


\section{Conclusion}
\label{sec:conclusions}
We introduce a multisensor pretraining model that leverages a novel cross-sensor paradigm to facilitate joint representation learning. This approach adeptly captures the intricate relationships between corresponding sensors. Built on a comprehensive multisensor dataset of over 2 million images, our model showcases outstanding performance across a variety of multisensor downstream tasks.

Looking ahead, there's potential to augment the model’s utility for downstream tasks where temporal information plays a pivotal role, such as in predicting ecosystem changes. The integration of temporal data holds immense value but introduces considerable challenges, primarily due to the significant increase in pre-training costs associated with incorporating temporal elements. Thus, the effective amalgamation of spatial and temporal information into a pretrained model demands more than just the inclusion of data; it necessitates a profound rethinking of methodological design.

{
    \small
    \bibliographystyle{ieeenat_fullname}
    \bibliography{main}
}

\clearpage

\appendix

\section{Optimization on Pretraining Data}

\subsection{Performance with other choices of pretraining data. } 
\label{pretraining_data}

To optimize \datasetname, we initially focused on optimizing \datasetname-RGB. As previous research has indicated, a successful pretraining dataset requires rigorous testing of each component \citep{nguyen2023quality}. Thus, we conducted a series of experiments on each individual dataset. These experiments involved the use of ImageNet \citep{imagenet} with 3 million images, GeoLifeCLEF with 3.3 million images, and the Functional Map of the World (FMoW) \citep{fmow}. For FMoW, we segmented the dataset into tiles of size 384, leading to a total of 6 million images. This diverse selection of datasets allowed us to comprehensively test and optimize our pretraining approach for \datasetname.

To ensure other variables, such as the backbone architecture and pretraining methodologies, do not skew our results, we chose to employ the Swin-base \citep{swin} and committed to pretraining from scratch. In line with our aim for equitable comparison, we also adhered to the same seven downstream tasks as delineated in the previous report \citep{mendieta2023gfm}. The results are shown in Table \ref{tab:dataset_ablation_1} and Table \ref{tab:dataset_ablation_2}. This approach creates a consistent testing environment across all datasets, reducing the potential for bias or error. 

Interestingly, upon integrating the GeoLifeCLEF into our testing framework, we observed a downturn in performance on downstream tasks. This result signifies that not all datasets necessarily contribute to improved model performance, and their selection demands careful consideration.

Even though the addition of both the Functional Map of the World dataset and ImageNet gave rise to performance metrics that were commensurate with those achieved by \datasetname-RGB, these new dataset additions were not as efficient. The key reason for this inefficiency was the significantly larger size of the pretraining dataset, which introduced higher computational costs and longer processing times. This finding highlights the importance of carefully balancing dataset size and complexity with computational efficiency in the model training process.

\begin{table*}[h]

    \centering
    \setlength\tabcolsep{3.0pt} 
    \begin{tabular}{cccccc}
        \toprule
        Dataset & \# Image & OSCD (F1) & DSFIN (F1) & BEN 10\% & BEN 1\% \\
        \toprule
        GeoPile \citep{mendieta2023gfm} & 600K & \textbf{57.5} & 66.2 & 86.4 & 79.3 \\
        \datasetname-RGB & 1.7M & 57.1 & \textbf{70.4} & \textbf{86.8} & \textbf{79.6}\\
        \datasetname-RGB + ImageNet \citep{imagenet} & 3M & \textbf{57.5} & 69.2 & 86.4 & 79.5\\
        \datasetname-RGB + GeoLifeCLEF & 3.3M & 56.1 & 61.6 & 86.1 & 78.9 \\
        \datasetname-RGB + FMoW \citep{fmow} & 6M & 58.2 & 69.3 & 86.2 & 79.1 \\
        \bottomrule
    \end{tabular}
    \caption{Results of downstream tasks with different pretraining datasets: change detection and classification}
    \label{tab:dataset_ablation_1}
\end{table*}

\begin{table*}[h]

    \centering
    \setlength\tabcolsep{3.0pt} 
    \begin{tabular}{cccccc}
        \toprule
        Dataset & \# Image & WHU & Vai. & SN2 (PSNR) & SN2 (SSIM)\\
        \toprule
        GeoPile \citep{mendieta2023gfm} & 600K & 90.1 & 75.1 & \textbf{22.626} & 0.645\\
        \datasetname-RGB & 1.7M & \textbf{90.6} & 75.9 & 22.599 & \textbf{0.658}\\
        \datasetname-RGB + ImageNet \citep{imagenet} & 3M & 90.5 & \textbf{76.1} & 22.107 & 0.631\\
        \datasetname-RGB + GeoLifeCLEF& 3.3M & 89.1 & 74 & 16.663 & 0.512\\
        \datasetname-RGB + FMoW \citep{fmow} & 6M & 90.2 & 75.7 & 22.448 & 0.638\\
        
        \bottomrule
    \end{tabular}
    \caption{Results of downstream tasks with different pretraining datasets: segmentation and super-resolution}
    \label{tab:dataset_ablation_2}
\end{table*}

\subsection{Performance without RGB modality}
\label{wo_RGB_modality}

RGB modalities are singled out because of the abundance of RGB datasets that come from various sources beyond just Sentinel-2. For instance, MillionAID \citep{Long2021DiRS}, a dataset comprised of a wide range of RGB images, is sourced from multiple satellites, including GeoEye, WorldView, QuickBird, IKONOS, and SPOT satellites, among others. Additionally, a previous study \citep{satmae} found that using only Sentinel-2 data for pretraining does not yield optimal performance in the downstream evaluation. Therefore, we sought to diversify our sources and include a wider range of RGB images in our pretraining data. This breadth of data sources significantly enriches the diversity of the RGB modality in our study. 

Despite overlapping GSD in some RGB modality, more geospatial features will be included. Although these datasets may not provide an imaging spectrum as wide as Sentinel-2, they enhance the entropy of pre-training data, which has been proven to be effective in \citep{mendieta2023gfm}, which is demonstrated by Table \ref{tab:RGB_modality}. 

\begin{table}[h]
    
    \centering
    \setlength\tabcolsep{3.0pt} 
    \begin{tabular}{ccc}
        \toprule
        Pretraining sensor modality & 10\% BEN & cloud removal\\
        \toprule
        Metric  &  mAP ($\uparrow$) & SAM ($\downarrow$))\\
        \toprule
        SAR (in Figure 3) & 84.2  &  8.37 $\pm$ 0.057\\
        Sentinel-2 (in Figure 3)  & 86.6 & 8.33 $\pm$ 0.034 \\
        RGB & 86.4 & 10.45 $\pm$ 0.12\\
        w/o RGB  & 86.6 & 9.67 $\pm$ 0.12\\
        \datasetname  & \textbf{87.7} & \textbf{7.51} $\pm$ 0.057\\
        \bottomrule
    \end{tabular}
    \caption{Results pretrained with single modality or without RGB modality.}
    \label{tab:RGB_modality}
\end{table}

\section{Pretraining Details}

\subsection{Pretraining Settings}
\label{pretraining_settings}

\textbf{Masking.} All hyper-parameters are listed in Table \ref{table:xgeo_optimization}. We implement a masking strategy that maintains consistency around different channels within the same sensor, applying the mask at the same locations. However, when it comes to different sensors, we employ a varying masking approach, ensuring that the mask is applied at different locations. This methodology allows us to preserve sensor-specific information while investigating inter-sensor discrepancies effectively. 

\begin{table*}[!tb]
    \centering
    \begin{tabular}{l|c}
    	\toprule[1.5pt]
    	Hyper-parameter & Value \\
        \midrule
        Image size & $192 \times 192$ \\
        \midrule
        Optimizer & AdamW \\
        \midrule
        $\beta_1$                   & $0.9$     \\
        \midrule
        $\beta_2$                   & $0.999$   \\
        \midrule
        Eps & $1.0\times10^{-8}$ \\
        \midrule
        Momentum & 0.9 \\
        \midrule
        Weight decay                & 0.05 \\
        \midrule
        Learning rate               & \{$1.0\times10^{-4}$, $0.25\times10^{-4}$, $1.0\times10^{-5}$\} for RGB, \\ & Sen12MS \citep{sen12ms} and MDAS \citep{MDAS}\\
        \midrule
        Warm up learning rate & $5.0\times10^{-7}$\\
        \midrule
        Weight decay                & $10^{-5}$ \\
        \midrule
        Batch size                  & \{$128$, $32$, $12$\} per GPU for RGB, \\ & Sen12MS \citep{sen12ms} and MDAS \citep{MDAS}\\
        \midrule
        Training epochs             & $800$ or $100$   \\
        \midrule
        Warm up epochs          & 10   \\
        \midrule
        Learning rate decay         & Multistep    \\
        \midrule
        Gamma & 0.1\\
        \midrule
        Multisteps & [700,] for 800 or [] for 100\\
        \midrule
        \# Experts              & 8 \\
        \midrule
        MoE blocks              & 1, 3, 5, 7, 9, 11, 13, 15, 17 (Every other swin block) \\
        \midrule
        Top-value ($k$) & 1 \\
        \midrule
        Capacity factor & 1.25 \\
        \midrule
        Aux loss weight ($\lambda$) & 0.01 \\
        \midrule
        Mask patch size & 32 \\
        \midrule
        Mask ratio & 0.6 \\
        \bottomrule[1.5pt]
    \end{tabular}
     \caption{Hyperparameters of \modelname pretraining.}
    \label{table:xgeo_optimization}
\end{table*}

\textbf{Heterogeneous batch size.} Given the disparity in the number of images obtained from different sensors, we employ a heterogeneous batch size strategy for our training process. This methodology adjusts the batch size in proportion to the amount of data sourced from each individual sensor. In essence, during each epoch of our training process, every type of sensor is iterated through once, irrespective of the data volume associated with that particular sensor. This ensures that all sensor types have an equal chance to contribute to the model's learning process, fostering a more balanced and comprehensive training regimen. Alongside this, we also adjust the learning rate proportionally in accordance with the batch size allocated per sensor.

\begin{figure}[h!]
\centering
    \includegraphics[width=0.5\textwidth]{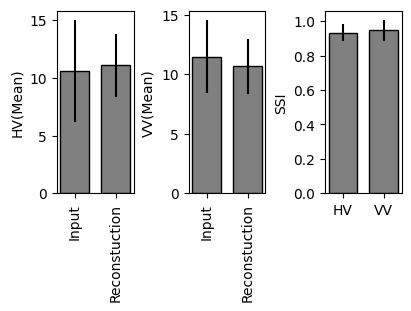}
\caption{Figure R1: SAR backscatter statistics comparing input and reconstruction using the MIM. The two bands of SAR are HV and VV. The mean and standard deviation for the HV band are shown on the left, while those for the VV band are displayed on the right. The Speckle Suppression Index (SSI) values are presented in the right panel. An SSI value closer to one indicates that the mean and standard deviation remain consistent before and after reconstruction.}
\label{fig:SAR_backscatter}
\end{figure}


\section{Downstream Experiments}

\subsection{Model size}
\label{model_size}

Regarding the number of parameters, we followed a standard backbone for pretraining, the details of which have been reported in \cite{simmim}. Comparisons between training from scratch and using ImageNet pretrained weights have been provided in Table \ref{tab:model_size} and corroborated by previous studies \citep{seco, satmae, mocov2, dinomc, mendieta2023gfm}. 

\begin{table*}[h]
    \centering
    \begin{tabular}{c|c|c|c|c|c|c|c|c|c}
        \toprule
        Model & SeCo & SatMAE & MoCoV2 & DINO-MC & GFM & \modelname\\
        \toprule
        \# of trainable parameters & 23M & 307M & 23M & 48.6M & 89M & 89M \\
        \midrule
    \end{tabular}
    \caption{Model size}
    \label{tab:model_size}
\end{table*}

\subsection{Experimental settings}
\label{experimental_settings}

\begin{figure}[h]
\centering
    \includegraphics[width=0.5\textwidth]{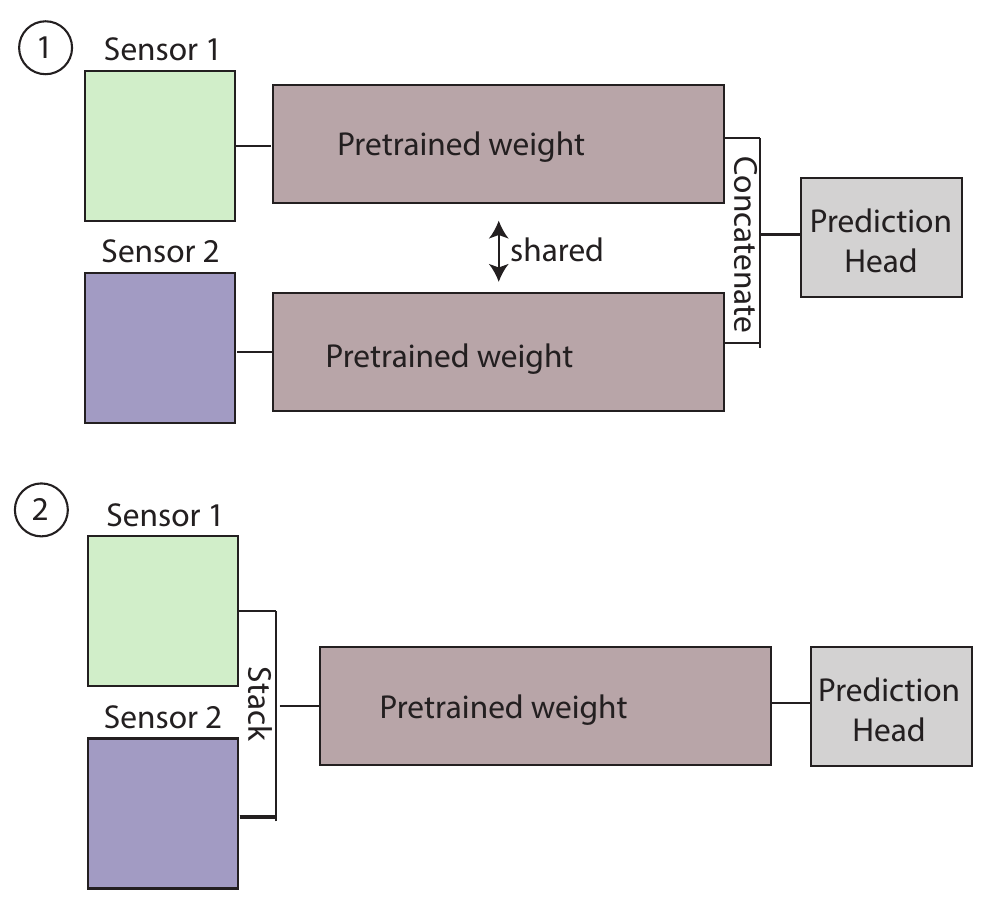}
    \caption{Two methods of downstream transfer. 
    In the top panel, every sensor is fed into a separate encoder initialized with \modelname pretrained weight. The embeddings from the last layer are concatenated, and then fed through the prediction head, such as classifier and segmentation decoder. In lower panel, images are concatenated along the color channel and then fed through one encoder initialized with \modelname pretrained weight. 
    }
    \label{fig:downstream_transfers}
\end{figure}

There are primarily two ways to leverage pretrained weights, as depicted in Figure \ref{fig:downstream_transfers}. The first approach involves feeding each sensor through encoders that share weights. The resulting embeddings are then concatenated and fed into the classifier. In the second approach, all sensor data are stacked together in the color channel prior to patchification. This approach resembles the multiMAE method \cite{multimae}, where the projected patches from all modalities are concatenated into a single sequence. Our experiments on 1\% of the BEN dataset \cite{BEN}, listed in Table \ref{tab:BEN_transfer}, demonstrate that both methods yield comparable results. However, the latter approach is more computationally efficient, meaning that the former approach takes longer time to reach the optimal performance and consumes more memory as well. Therefore, all results mentioned in the main text utilize this second approach. Importantly, in both cases, no masking is performed during the transfer phase.

\begin{table}[h]

    \centering
    \setlength\tabcolsep{3.0pt} 
    \begin{tabular}{cc}
        \toprule
        Finetuning Method & BEN 1\% \\
        \toprule
        1 & 80.8\\
        2 & 80.8 \\
        \bottomrule
    \end{tabular}
    \caption{Results of BEN when comparing different downstream transfer methods illustrated in Figure \ref{fig:downstream_transfers}}
    \label{tab:BEN_transfer}
\end{table}



\subsection{Visualization}
\label{qualitative_examples}

\begin{figure}[h]
\centering
\includegraphics[width=0.5\textwidth]{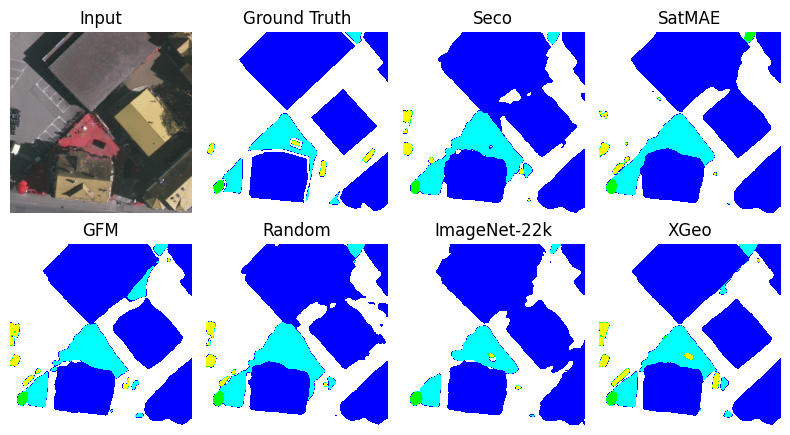}
\caption{A display of qualitative results showcasing segmentation outcomes from \modelname in comparison to other competitive methods.}
\label{fig:vaihingen}
\end{figure}

We present some quantitative results of segmentation in Figure \ref{fig:vaihingen} respectively. 

\end{document}